# Multi-Scale PCB Defect Detection with YOLOv8 Network Improved via Pruning and Lightweight Network

LI Pingzhen, XU Sheng**, CHEN Jing*, SU Chengyue

(School of Physics and Optoelectronic Engineering, Guangdong University of Technology, Guangzhou 510000, China)

## summaries

With the high density of printed circuit board (PCB) design and the high speed of production, the traditional PCB defect detection model is difficult to take into account the accuracy and computational cost, and cannot meet the requirements of high accuracy and real-time detection of tiny defects. Therefore, in this paper, a multi-scale PCB defect detection method is improved with YOLOv8 using a comprehensive strategy of tiny target sensitivity strategy, network lightweighting and adaptive pruning, which is able to improve the detection speed and accuracy by optimizing the backbone network, the neck network and the detection head, the loss function and the adaptive pruning rate. Firstly, a Ghost-HGNetv2 structure with fewer parameters is used in the backbone network, and multilevel features are used to extract image semantic features to discover accurate defects. Secondly, we integrate C2f-Faster with small number of parameters in the neck section to enhance the ability of multi-level feature fusion. Next, in the Head part, we design a new GCDetect detection head, which allows the prediction of bounding boxes and categories to share the weights of GroupConv, and uses a small number of grouping convolutions to accomplish the regression and classification tasks, which significantly reduces the number of parameters while maintaining the accuracy of detection. We also design the Inner-MPDIoU boundary loss function to improve the detection and localization of tiny targets. Finally, the model was pruned by an optimized adaptive pruning rate to further reduce the complexity of the model. Experimental results show that the model exhibits advantages in terms of accuracy and speed. On the publicly available PCB defect dataset, mAP0.5 reaches 99.32% and mAP0.5:0.9 reaches 75.18%, which is 10.13% higher compared to YOLOv8n. The amount of parameters is reduced to 0.67M, which is equivalent to 22% of YOLOv8n, the number of floating-point operations is reduced to 2.4G, which is equivalent to 22% of YOLOv8n, and the size of the model is also reduced from In terms of inference speed, the single image inference speed is improved from 119.1 to 129.3, and the 32-image inference speed is improved from 1030.1 to 1219.4, which provides an accurate and efficient detection method for PCB defect detection.

## introductory

PCB is one of the basic components of various electronic devices and plays a vital role in modern electronic devices, but the manufacturing process is susceptible to a variety of factors such as mechanical friction, electrostatic interference and chemical corrosion, etc. [1] . These factors may lead to a series of PCB defects such as short circuits, open circuits, scratches, and stray copper. These defects may lead to electronic equipment dysfunction, performance degradation, and even seriously affect the normal operation of the entire system, bringing potential safety hazards! [2] . Therefore, to ensure the quality and stability of PCBs, to ensure the normal operation of electronic devices and user experience, the detection of PCB defects is critical! [3]. .

Automatic optical defect detection methods, as a non-contact in-line inspection technology, have become an important tool for quality control of printed circuit boards, and their core algorithms are broadly categorized into traditional machine vision inspection methods and deep learning-based methods. Machine vision inspection methods use traditional methods based on image processing, such as threshold segmentation [4] , edge detection and segmentation methods [5] , and BP neural networks and support vector machines (svm) [6]. [6] Traditional machine learning methods represented by BP neural networks and support vector machines (svm) [6] and other techniques to realize PCB defect detection have solved the problem of automatic optical detection of PCBs to a certain extent, but there are drawbacks such as relying on hand-designed feature extraction, poor robustness, low adaptability, and limited accuracy, which are especially ineffective in dealing with complex backgrounds and tiny defects. In contrast, deep learning-based methods can automatically learn abstract features, have higher detection accuracy and strong adaptability, and can effectively deal with a variety of PCB defects, with significantly improved real-time and robustness, thus becoming a research hotspot in recent years! [7]. It has become a research hotspot in recent years [7].

Deep learning research in PCB defect detection algorithms is mainly divided into two categories: one is single-stage target detection networks, such as YOLO and SSD; the other is two-stage target detection networks, such as R-CNN, Faster R-CNN, etc. BING HU et al. improved the Faster R-CNN network by using ResNet50 as the backbone network, the ShuffleNetV2's residual cell structure to accelerate the network, while introducing GARPN to detect more accurate defective targets [8] Runwei Ding et al. proposed TDD-Net, a Faster R-CNN-based tiny defect detection network, using a feature pyramid structure and fusing multi-scale features for the task of small target defect detection [9]. [9] . Although the two-stage target detection method performs better in terms of accuracy, it needs to generate candidate regions first, and then classify and localize them, which leads to slower detection speed and is difficult to meet real-time requirements. In contrast, the single-stage target detection method combines target classification and localization into a single step, which reduces the computational complexity and significantly improves the detection speed, which is more suitable for real-time scenarios, and especially performs well in tasks that require fast and efficient detection.

On the study of single-stage target detection network for PCB defect detection, Wei Shi et al. proposed a single-shot detector (SSDT) with better robustness compared to the SSD, which enhances the recognition of targets in complex backgrounds by using an attention mechanism to learn the relationship between the features to be fused, and employing a shuffle module to avoid the effect of aliasing after the features have been fused [10] Chen's YOLOX-based Attention Mechanism and Multiple Feature Fusion (AM FF-YOLOX) method integrates the attention mechanism and adaptive spatial feature fusion in the feature extraction network to enhance the attention to the detected objects [11]. [11] Liu et al. explored the Gaussian intersection of union (GsIoU) loss function based on YOLOv4 and applied it to PCB component detection by optimizing the accuracy of box regression [12]. [12]. Liu et al.Although single-stage target detection networks have improved in accuracy, their complex structure leads to high resource consumption and long operation time [13] . Complex models usually require more computation and memory, which makes deployment in real-time applications limited. To cope with these problems, lightweight models have emerged. By reducing the number of parameters and the amount of computation, lightweight models dramatically reduce resource requirements and latency while maintaining performance, making them more suitable for edge devices and real-time detection tasks. In this regard, YOLO series models have been further improved by integrating various new features, including the attention mechanism [14], [15], [16] and clustering algorithms [17], [18], [19], [20]. [17],

[18] The modified loss function [17], [18] , and model size reduction by using different convolution operations.Jiansheng Liu et al. proposed Mixed YOLOv4-LITE, a lightweight YOLOv4-based defect detection framework, which uses MobileNet and depth-separable convolution to replace the backbone network and the traditional convolution, and improves the speed of detection with a small amount of accuracy loss [19] Tang, JL proposed a YOLOv5-based PCB defect detection algorithm PCB- yolo [20] Zhao, Q introduced ShuffleNetV2 into YOLOv5 [21]. [21] to further improve the accuracy of defect detection. du, BW et al. replaced the modules in YOLOv5s network by adding a moving inverted bottleneck convolution (MBConv) module, a convolutional block attention module (CBAM), a bi-directional feature pyramid network (BiFPN), and a deep convolution, thus enhancing the model's feature extraction capability and information fusion capability [22] .

The existing PCB defect detection based on the YOLO series of algorithms has achieved relatively good results, but the accuracy and real-time performance of the detection of tiny defects in complex backgrounds need to be further improved, and in production, tiny defect points only account for a small portion of the image of printed circuit boards [23] In production, microdefects represent only a small portion of the printed circuit board image [23], which is characterized by random locations and minimal deviation from the background information, leading to the loss of critical information in PCB microdefect detection [24]. [24] This results in the loss of critical information in PCB micro-defect detection [24], which leads to imperfect extraction of micro-defect feature information and reduces the accuracy of detection. In addition, the current PCB production line is becoming more and more high-speed, which requires faster and faster detection speeds , so it is necessary to pay special attention to the tiny features, reduce the loss of critical information, and accelerate the detection process to adapt to the needs of high-speed PCB production lines.

Based on the shortcomings in the above studies, this paper proposes a lightweight PCB defect detection network based on YOLOv8, which achieves a satisfactory balance in accuracy, model size and detection speed. The main contributions of this paper are as follows:

1. An improved multi-scale PCB defect detection framework based on YOLOv8 is proposed with several structural innovations. First, the improved Ghost-HGNetv2 is used as the backbone network, to reduce the feature information loss, and to enhance the multilevel feature extraction capability with less number of parameters in the model. Secondly, Faster-Block is used to replace the traditional C2f Bottleneck structure to improve the feature fusion capability. Finally, a new type of GCDetect detection head is designed, the prediction of bounding boxes and categories share the weights of GroupConv, and the regression and classification tasks can be efficiently accomplished by using a small number of grouping convolutions, and the above measures significantly improve the detection speed while maintaining high accuracy.

2. During the training process, the Inner-MPDIoU boundary loss function is designed to enhance the detection and localization of tiny targets, combining the advantages of Inner-IoU detection and MPDIoU localization. , considering the overlapping region, the distance from the center point, and the deviation of the width and height, the auxiliary border of the tiny target is controlled using the scale factor ratio to fit the boundary of the tiny target more accurately, so as to improve the localization accuracy of the tiny target.

3. Using adaptive channel pruning algorithm on the trained model, the pruning rate was optimized, and the optimal pruning rate was investigated to reduce the number of parameters and computation of the model while maintaining the performance of the model, and to further improve the speed of model detection.

**The rest of the paper is organized as follows:** Section II reviews the application of multiscale feature fusion techniques in small target detection and PCB defect detection methods. Section III details the proposed model improvement strategy. Section IV presents the experiments, including the description of the dataset and the process of model training. Section V presents the result analysis, which validates the proposed algorithm and compares its performance with other algorithms. Finally, Section VI is the conclusion, which summarizes the main findings of this paper and future research directions.

# Related work

A、Multi-scale feature fusion for small target detection

Multi-scale feature fusion enhances the model's ability to detect small targets by combining different levels of feature maps [25] . Small targets are difficult to detect in images due to their small size and inconspicuous features, resulting in difficult detection, especially in complex backgrounds. To address this challenge, researchers have proposed a variety of deep learning-based approaches that typically utilize the hierarchical feature extraction capability of convolutional neural networks (CNNs) to enhance the model's ability to detect small targets through feature fusion strategies.

Existing multiscale feature fusion methods are mainly classified into two main categories: feature pyramid-based fusion strategies, such as feature pyramid networks (FPNs) and their variants, which achieve multiscale feature fusion through top-down paths and lateral connections; and feature fusion augmented by attentional mechanisms, such as CBAMs, which, by focusing on important features and suppressing unimportant information to improve the detection performance [26] Chen et al. (2022)2 proposed a four-pillar strategy, including multi-scale representation, contextual information, super-resolution, and region suggestion, which significantly improves small target detection performance through multi-scale feature fusion [27] Bosquet et al. (2020) proposed STDnet to enhance small target detection by utilizing fused multiresolution feature maps [28] Cao et al.'s (2017) FFSSD improves small target detection by fusing features at different scales at a shallower level [29] . While Duan et al.'s (2020) CADNet enhances feature representation of small targets by channel-aware deconvolutional network [30] Tang et al. (2023) added a small-target detection layer to the feature fusion network of YOLOv5 for small-size defects in PCBs, which enhanced the detection of small targets, but introduced too many parameters [20] Wang et al. (2022) proposed a small target defect detection algorithm for steel surfaces based on improved YOLOv7, which used a de-weighted bidirectional feature fusion network to further fuse multiscale features [31] Li, Zhao et al. (2022) proposed an industrial small defect detection method based on extended perceptual domain and feature fusion by employing a new multiscale weighting strategy in the feature fusion network of YOLOv5s [32] .

However, there are still some shortcomings of these methods for PCB tiny defect detection. First, with the increase of network depth, the feature map may lose important detail information during the transmission process, and these are unfavorable for small target detection. Second, many methods enhance the feature fusion effect by introducing a complex attention mechanism or Transformer module, which inevitably increases the computational complexity of the model and the number of parameters, leading to a decrease in the inference speed. Small target detection in complex contexts still faces greater challenges because these defects have less feature information and more feature interference. We adopt Ghost-HGNetv2 for multiscale fusion, which does not introduce a complex attention mechanism or Transformer module, but fuses multiscale features during multiple downsampling and upsampling, reducing the network complexity and number of parameters. In order to cope with fewer de fective features and background interference, a recursive structure is used to allow information to be passed and fed back multiple times, which promotes the exchange between local and global information and enhances the model's ability to learn complex patterns. In order to reduce the detail information lost in the passing process of feature maps, Ghost-HGNetv2 uses skip connections during downsampling and upsampling, which enhances the model's ability to perceive fine structures and thus reduces information loss.

B. Lightweighting of small target detection networks

Current work on lightweighting of small target detection networks encompasses methods such as depth separable convolution, channel pruning, network pruning, knowledge distillation, lightweight model design, and lightweight modules. Deeply separable convolution decomposes the standard convolution operation into two steps, deep convolution and point-by-point convolution, which significantly reduces the computational effort. Howard et al. proposed an efficient lightweight model, MobileNetV3, which achieves high-performance small-target detection by means of deeply separable convolution and Neural Architecture Search (NAS) techniques [33] Alessandro Betti et al. used the idea of depth separable convolution to optimize the network in order to improve the efficiency of the YOLO-S model for small target detection in aerial images and to achieve a model lightweighting [34]. The The lightweight model design focuses on constructing a new network structure, optimizing the model in terms of the overall architecture, so that it has fewer parameters while guaranteeing the performance of small target detection.Yuwen Li et al. redesigned the network by introducing the SSFF module, improving the feature pyramid network, and so on, in order to achieve the purpose of lightweighting and enhancing the ability of small target detection [35]. [35] . Introducing lightweight modules to replace or add some modules in the existing network structure to reduce the volume, without adding too much computational burden, Rujin Yang et al. used the GhostNet module to replace the backbone network and the correlation convolution in the neck in the original algorithm of YOLOv5, to reduce the number of parameters and to realize lightweighting [36] . Channel pruning aims to remove unimportant channels in the network, thus reducing the number of parameters and computation of the model. Huang Linglin et al. optimized the traditional YOLO network by pruning and fusing multiple features, reducing the computation and improving the ability to detect small targets in real time [37] . Network pruning reduces the number of parameters by cutting out unimportant connections or neurons in the network without significantly degrading the performance of the model, and Golnaz Ghaisi et al. were able to reduce the computational burden while maintaining the accuracy by adaptively adjusting the network architecture to different sizes of target detection tasks [38]. [38] . In this paper, a comprehensive strategy was used for the above approaches, the whole network was designed to be lightweight, the backbone network, the neck network and the detection head were lightweighted, and after the optimization of the whole network, it was not suitable for pruning the whole network again, so adaptive pruning method was used and the optimization of the pruning rate was searched for and the channel pruning was carried out to further reduce the complexity of their model for deployment on the edge devices.

# methodologies

In this paper, various improvements are made based on the YOLOv8 network, keeping the main architecture of the model unchanged, i.e., it consists of three parts: the backbone network, the neck network and the head network. The main improvements in this paper include:

Backbone network optimization: We improve Ghost-HGNetv2 as the backbone network, to reduce the feature information loss, to adopt a multi-level feature extraction strategy, which effectively combines the low-level local features with the high-level global semantic information, to generate a more complete feature map to improve the accuracy of target detection. By introducing the GhostConv module, it also has high parameter efficiency and computational efficiency.

Neck network optimization: the neck network is responsible for feature fusion and further processing in the target detection model, which directly affects the detection performance and efficiency of the model. In order to optimize the feature fusion process, we introduce the C2f-Faster module in the neck network. The clever design of the C2f-Faster module improves the efficiency of multi-scale feature fusion, which not only enhances the model's ability to perceive small targets, but also effectively reduces the complexity of the model.

Detection head optimization: the detection head, as a key component of the target detection model, directly determines the model's ability to identify and localize targets. In order to improve the detection accuracy and speed, we design and adopt GCDetect as the new target detection head, which uses GroupConv instead of the traditional convolution operation, and adopts the strategy of sharing the weights of GroupConv between the bounding box (Bbox) and the categorization (Cls), which significantly reduces the number of parameters and the amount of computation in the model, and increases the detection speed, and also enhances the adaptability of the model to different scales of targets, which makes the model more adaptable to complex scenes. detection speed, and also enhances the model's ability to adapt to targets of different scales, which guarantees the application of the model in complex scenes.

Loss function optimization: the loss function plays a crucial role in the training process of the target detection model, which directly affects the convergence speed and detection accuracy of the model. In order to further optimize the training effect of the model, we adopt Inner-MPDIoU to replace the original CIoU loss function. Inner-MPDIoU integrates the detection of Inner-IoU and the localization advantages of MPDIoU, takes into account the overlapping area, the distance from the center point, and the deviation of width and height, and controls the generation of the auxiliary borders of small targets through the introduction of a scale factor ratio, thus accelerating the model convergence speed and improving the accuracy. the auxiliary borders of small targets by introducing the scale factor ratio, which accelerates the convergence speed of the model and improves the accuracy of detection. This optimization strategy makes the model perform better in small target detection, and also enhances the robustness and generalization ability.

Model Pruning: after the training of the target detection model is completed, in order to further improve the speed and lightness of the model, we use an adaptive structured pruning based on the LAMP algorithm to prune the channels of the model. This pruning strategy reduces the number of parameters and computation of the model by removing channels that contribute less to the model, while maintaining the performance of the model. We compared different pruning rates and chose the optimal pruning rate, which significantly improves the detection speed of the model and the efficiency of practical applications.

With the above improvements, the lightweight PCB defect detection network based on the improved YOLOv8 proposed in this paper achieves a better balance in terms of accuracy, model volume and detection speed. This balance improvement solves the performance bottleneck that is difficult to be balanced in existing research, thus providing an effective solution for PCB quality control in real industrial applications. By optimizing the backbone network, the neck network and the detection head, as well as adopting more refined loss functions and model pruning techniques, our improvements allow the detection network to improve in accuracy while substantially compressing the model size, further increasing the detection speed. The overall architecture of the improved network is shown in Fig. 1.

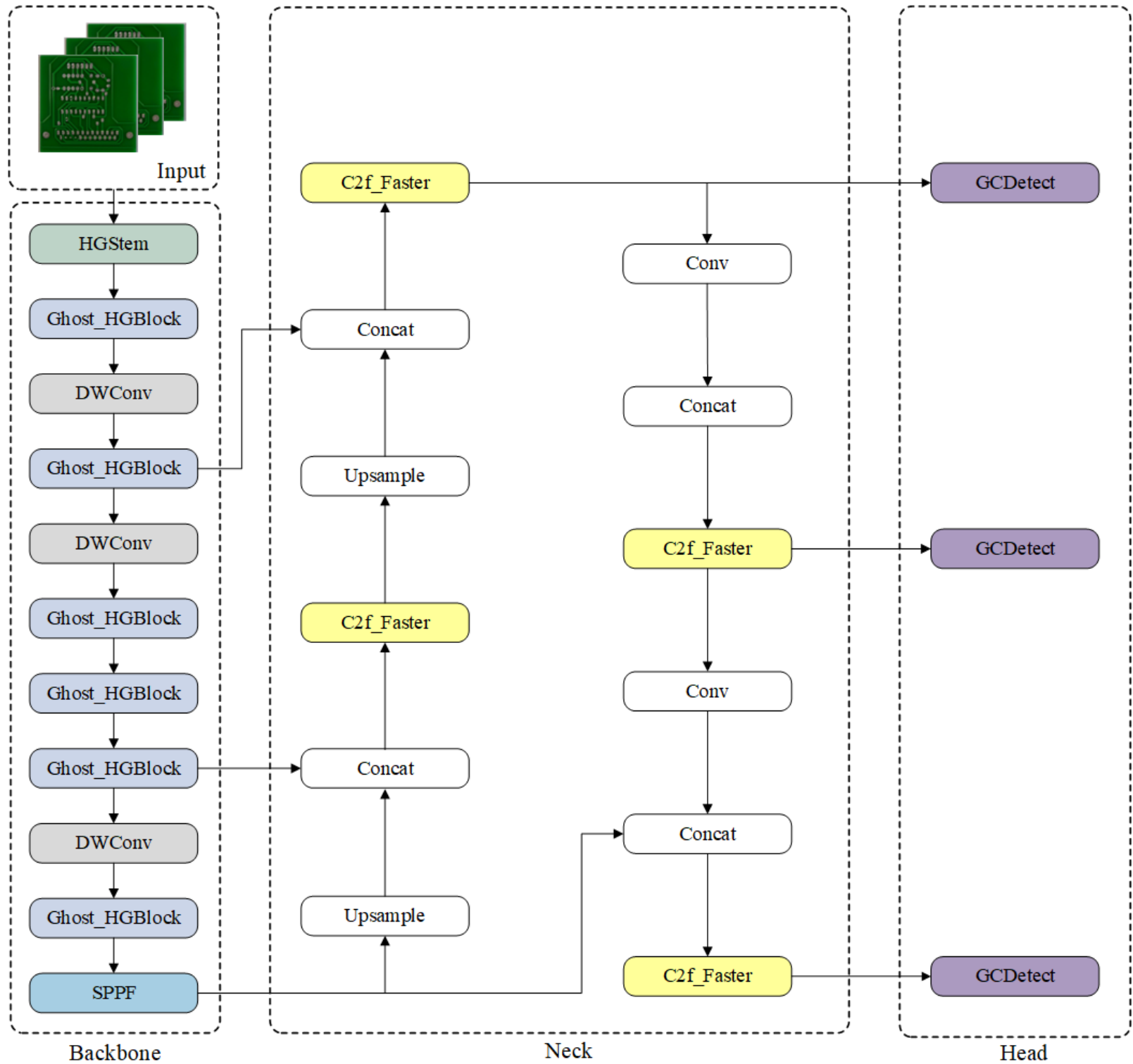

Figure 1. Structure of improved YOLOv8 network.

A. Backbone: Replacing Backbone with Ghost-HGNetV2 Architecture

In optimizing the YOLOv8 model, we propose to replace its backbone network with Ghost-HGNetV2 to improve network performance and reduce computational redundancy. The backbone network in the original YOLOv8 model contains multiple regular convolutional layers (Conv) and C2f modules, which fuse contextual information through the SPPF module [39] . However, conventional convolutional layers often produce redundant feature maps during feature extraction, and this redundancy increases the computational complexity, the number of model parameters, and the amount of floating-point operations, leading to slower inference and increased resource consumption. In order to optimize these problems, we replace the original backbone network with the optimized Ghost-HGNetV2 and improve it, aiming to reduce the feature information loss, enhance the feature extraction capability, and improve the detection performance by reducing the redundant computation, decreasing the number of model parameters, and speeding up the inference.

Ghost-HGNetV2 is able to efficiently capture both local details and global semantic information of an image at different scales through a multi-level feature extraction strategy, especially in small target detection and target differentiation in complex backgrounds, which demonstrates obvious advantages. Compared with the original network, our Ghost-HGNetV2 makes multiple improvements: first, the HGStem module in the [40] First, in the HGStem module [40], we adopt a smaller 2×2 convolutional kernel to extract local features, and improve the inference speed of target detection by reducing the number of channels. Second, by replacing the standard convolution with the PW+DW 5×5 combined convolution, we not only reduce the number of parameters in the model, but also expand the receptive field, thus improving the detection accuracy.

In addition, the GhostConv module is introduced to replace the traditional HG_Block, which reduces redundant features by decomposing the convolution operation, which significantly improves the inference speed and reduces the computational complexity. GhostConv generates the feature maps by decomposing the standard convolution into two parts: the first part generates the feature maps through the standard convolution; and the second part generates the "Ghost" features through the inexpensive linear transformation of the The second part generates the "Ghost" features by an inexpensive linear transformation, thus reducing the amount of computation. Its mathematical expression can be expressed as equation 1:

$$\text{GhostConv}(X) = \bigcup_{i=1}^{k} \Phi_i(X) \quad \#(1)$$

where $X$ is the input feature map, and $\Phi_1(X)$ . $\Phi_2(X)$ ... $\Phi_k(X)$ are the feature maps generated by different convolution operations, and the sy mbol ∪ denotes the stitching operation of the feature maps.

Ghost_HGBlock is significantly optimized in several ways, especially in the GhostConv module and recursive feature handling. We introduce equation (2) to describe this process:

$$\mathbf{F}_{out} = \mathcal{F}_{conv}\left(\bigoplus_{i=1}^{3} \mathcal{G}^i(\mathbf{F}_{in})\right) \oplus \mathbf{F}_{in} \quad \#(2)$$

In the Ghost_HGBlock module, the input feature map $\mathbf{F}_{in}$ After three recursive GhostConv operations $\mathcal{G}^i$ (i ∈ {1, 2, 3} denotes the number of recursive layers) to gradually extract feature information at different levels. The GhostConv operation at each layer reduces redundancy and improves computational efficiency by decomposing the convolution, which enhances the multi-scale representation of the feature graph and enables the network to capture both local details and global semantic information.

After three GhostConv operations, you get $\mathbf{F}_{ghost1}$, $\mathbf{F}_{ghost2}$, $\mathbf{F}_{ghost3}$, these feature maps are merged into a high-dimensional feature map by stitching operation $\bigoplus_{i=1}^{3}$ to merge into a high-dimensional feature map. Subsequently, after convolution operation $\mathcal{F}_{conv}$ fusion is performed to further extract useful feature information and enhance the ability to capture local and global semantics In order to preserve the spatial fine-grained information in the input feature map, jump-joining is performed to add $\mathbf{F}_{in}$ added to the convolved feature map to obtain the final output Fout. This mechanism effectively avoids information loss and conveys finer spatial details.

Ghost_HGBlock significantly improves inference speed with little or no impact on accuracy by optimizing convolutional operations, recursive structures, and jump connections. Compared with traditional CNN and Transformer architectures, Ghost-HGNetV2 has superior performance in small target detection and complex contexts, providing an efficient and lightweight backbone network for YOLOV8.The schematic diagram of HGStem and Ghost_HGBlock in Ghost-HGNetV2 is shown in Fig. 2.

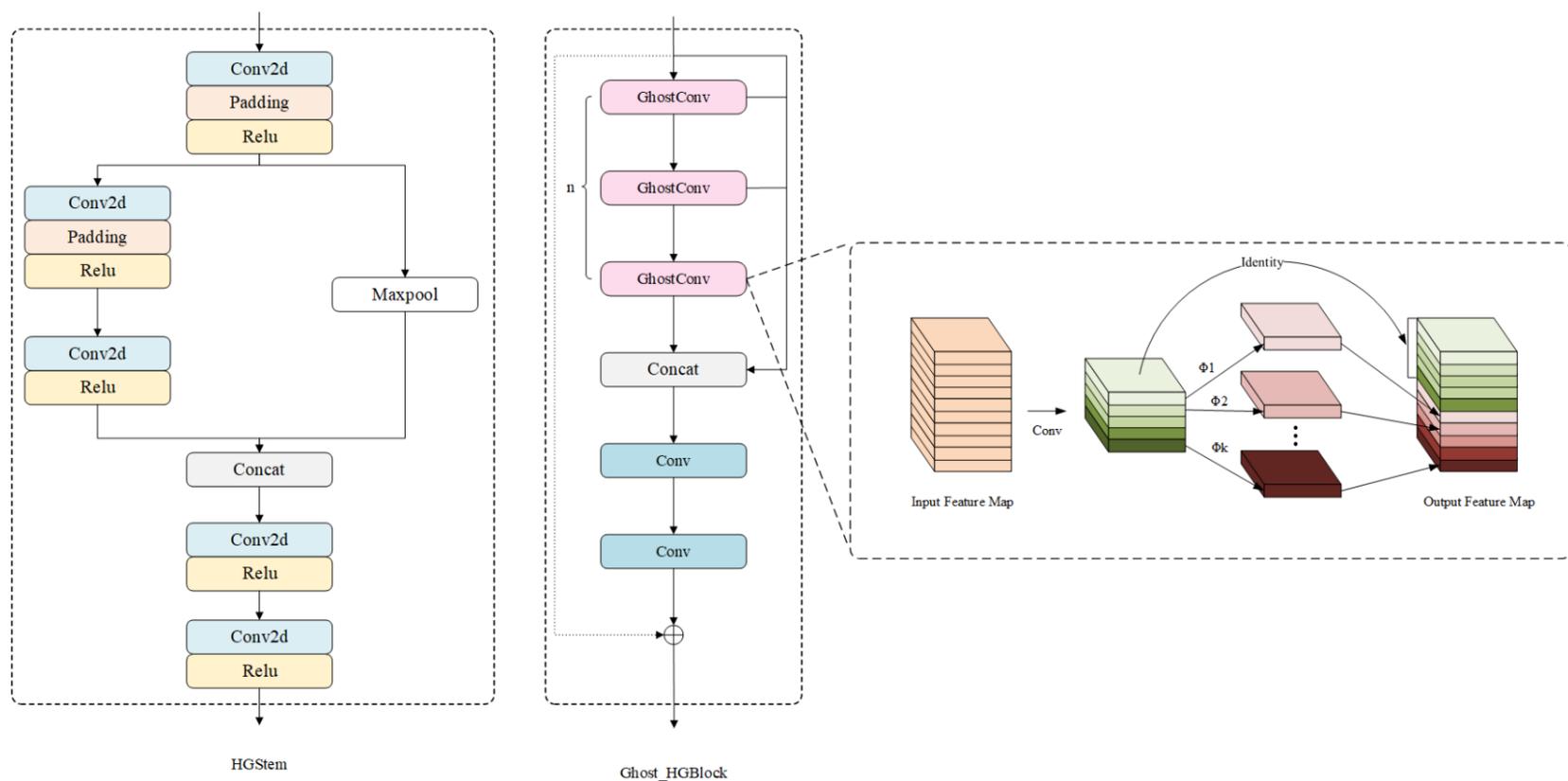

Figure 2. Ghost-HGNetv2 Network Architecture Diagram

B. **Neck**: Integrating C2f-Faster Module to Enhance Neck Features

In the optimization of the neck network of YOLOv8, the C2f-Faster module was introduced to enhance the multi-scale feature fusion capability to significantly improve the detection performance of the model.The design of the C2f-Faster module is inspired by FasterNet, which utilizes an efficient partial convolution (PConv) strategy for feature extraction [41] . In the traditional YOLOv8 architecture, the neck network realizes feature map fusion through multiple C2f modules. However, the bottleneck (Bottleneck) structure in the C2f modules contains two 3×3 convolution operations, which not only increases the number of parameters and computational overhead of the model, but also may introduce redundant features, which in turn reduces the computational efficiency.

To solve this problem, the C2f-Faster module introduces the FasterBlock module as a core component that generates output feature maps through a composite operation $X_{out}$. This process first uses local convolution (PConv) for feature extraction and then combines it with the standard convolution (Conv) operation to generate the final output feature map. The specific process can be represented as:

$$X_{out} = X + \text{Conv}(\text{PConv}(X)) = X + \sum_{j=1}^{C_p} W_j^{conv} * \left(\sum_{i=1}^{C} W_i \cdot X_i + b\right) \quad \#(3)$$

where $X$ is the input feature map, and $W_j^{conv}$ denotes the weights in the standard convolution, and $W_i$ is the weight in the local convolution operation, and $b$ is the bias term. By this combination, the details of the feature map are finely extracted, and at the same time, the expressiveness of the feature map is enhanced, providing richer feature information for the subsequent processing stage.

PConv does this by representing the input feature map as a three-dimensional tensor (h × w × c), where h, w, and c are the height, width, and number of channels of the feature map, respectively. In the convolution operation, PConv only convolves a quarter of the channels of the input features (cp = c/4), while retaining the rest of the features through residual concatenation, which effectively reduces the amount of computation and ensures the smooth transfer of information in the network. Since PConv only convolves a portion of the channels, its computational complexity is significantly lower than that of conventional convolution. h × w × k² × cp², which is 1/16th of conventional convolution, significantly reduces the computational burden and improves the model efficiency.

C2f-Faster In the feature fusion phase, the input feature map $X \in \mathbb{R}^{H \times W \times C}$ is convolved to obtain the output feature map $X_{conv}$, which is then split into two parts: $X_{split1}$ and $X_{split2}$. Subsequently $X_{split1}$ will be input to n FasterBlock one by one for processing, and the output is $Y_n$, whose process is shown in equation (4):

$$Y_n = \prod_{i=1}^{n} \text{FasterBlock}_i(X_{\text{split1}}) \quad \#(4)$$

It is worth noting that initially $Y = X_{\text{split1}}$ i.e. $X_{\text{split1}}$ is used as the input to the first FasterBlock. This operation allows the feature map to be processed in depth at multiple levels and angles, which helps the model to better capture complex visual patterns.

In the channel connection step, connect the $X_{\text{split2}}$ with the output of the last FasterBlock $Y_n$ is spliced to form a new feature map $X_{\text{concat}}$ that is shown in equation (5):

$$X_{\text{concat}} = \text{Concat}(X_{\text{split1}}, Y_n) \quad \#(5)$$

Finally, the spliced feature map Xconcat is fused with feature information by convolution operation to generate the final output feature map Xout, as shown in equation (6):

$$X_{\text{out}} = \text{Conv}(X_{\text{concat}}) \quad \#(6)$$

By introducing the C2f-Faster module, the model has significantly improved in detection performance and accuracy while the computation amount has been effectively reduced. The number of parameters of the model is reduced by about 14.72%, and the computation amount is reduced by about 10.29%, thus realizing the effective saving of computational resources. The module structure diagram of C2f-Faster is shown in Fig. 3.

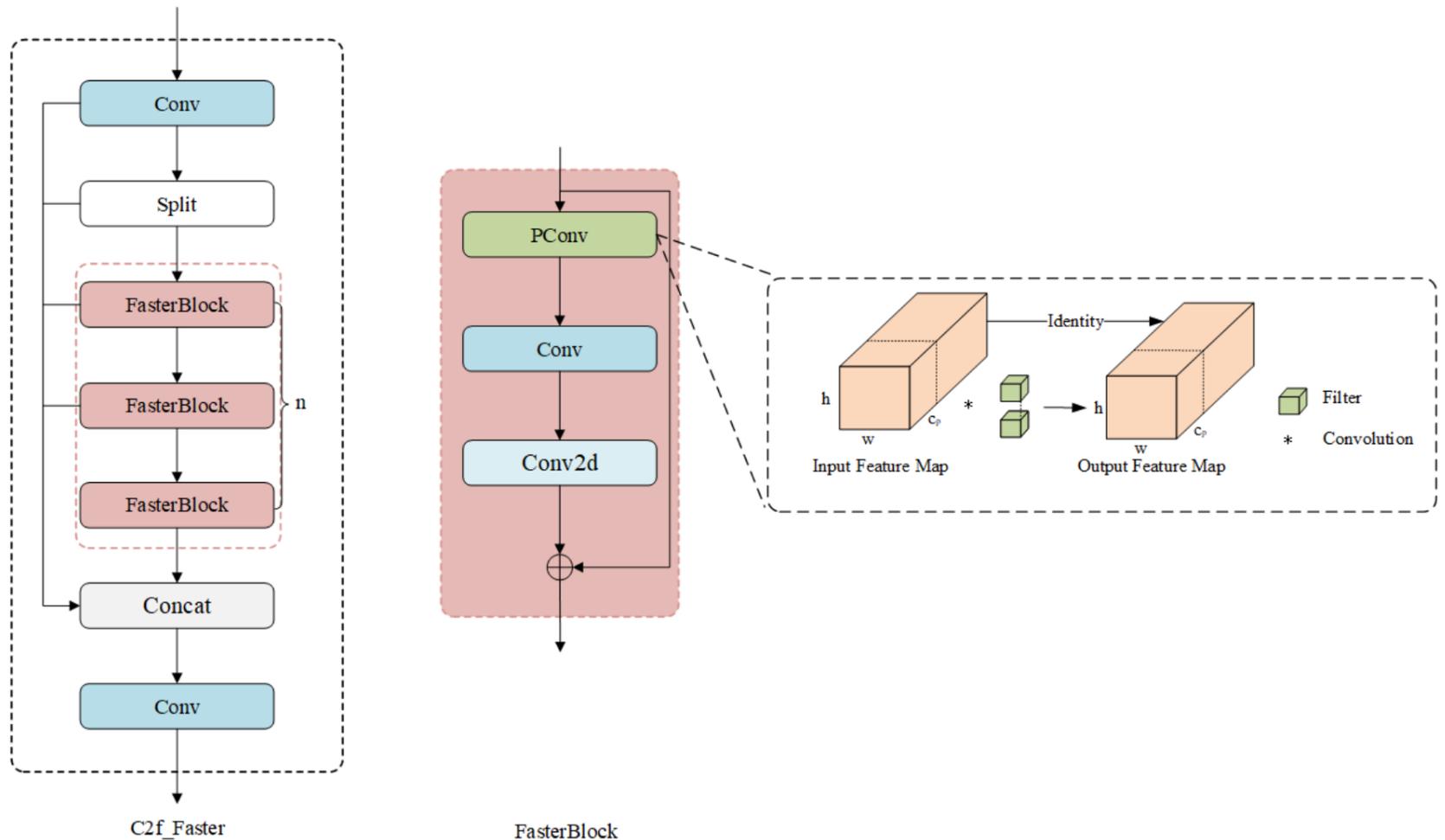

Figure 3. C2f-Faster Module Structure Diagram

**C. Head: Designing an Efficient Lightweight Detection Head by** GroupConv

In the original design of YOLOv8, the neck network generates three different scales of feature maps, each corresponding to a different receptive field, for detecting small, medium, and large targets, respectively. To process these multi-scale feature maps, YOLOv8 employs three independent detection heads, each dedicated to one scale of features. However, the original detection head contains multiple conventional 3×3 convolutional layers, which leads to a significant increase in the number of model parameters, and thus a significant rise in computational complexity and memory footprint.

To solve these problems, we propose a new lightweight detection header, GCDetect, which combines the GroupConv technique. GroupConv effectively reduces the number of parameters by dividing the input feature maps into subgroups and performing convolution operations on each subgroup, while solving the memory bottleneck through the parallel computing power of GPUs. GCDetect uses two tandem 3×3 group convolution layers instead of multiple convolution layers in the original detection header, and at the same time, divides the feature maps into 16 groups, and shares these two group convolutional layers between the classification and regression tasks. two groups of convolutional layers between classification and regression tasks. This design significantly reduces the number of parameters and computations, and increases the detection speed while maintaining the accuracy and robustness of the model. The mathematical expression for group convolution is shown in equation (7):

$$\text{GroupConv}(X) = \bigoplus_{g=1}^{G} (W_g * X_g) \quad \#(7)$$

where $X \in \mathbb{R}^{H \times W \times C}$ denotes the input feature map, and $X_g \in \mathbb{R}^{H \times W \times C_g}$ The table is the feature map $X$ of the gth group of subsets, and $C_g = C/G$ (here G = 16). $W_g \in \mathbb{R}^{k \times k \times C_g}$ is the convolutional kernel applied to group g, with $k = 3$ is the size of the convolution kernel and G is the total number of groups. * denotes the convolution operation, and ⊕ denotes splicing the outputs of all groups.

In GCDetect, two tandem group convolutional layers drastically reduce the number of parameters while maintaining the spatial structure of the feature map, which ensures that the lightweight design does not affect the detection accuracy.

In addition to reducing parameters, GCDetect introduces a feature sharing mechanism between the classification and regression tasks, which is similar to the weight sharing method in YOLOX. By making the classification and regression tasks share the same set of convolutional layers, the complexity of the model is further reduced. The classification output and regression output are computed as in Eq. (8), Eq. (9):

$$\hat{Y}_{\text{Cls}} = \text{Conv2d}_{\text{Cls}}(\text{GroupConv}_2(\text{GroupConv}_1(X))) \quad \#(8)$$

$$\hat{Y}_{\text{Bbox}} = \text{Conv2d}_{\text{Bbox}}(\text{GroupConv}_2(\text{GroupConv}_1(X))) \quad \#(9)$$

where $\hat{Y}_{\text{Cls}}$ denotes the categorization output, indicating the predicted target category probability; $\hat{Y}_{\text{Bbox}}$ denotes the bounding box regression output, representing the predicted bounding box coordinates. $X \in \mathbb{R}^{H \times W \times C}$ is the input feature map shared by the classification and regression tasks, and $\text{GroupConv}_1$ and $\text{GroupConv}_2$ are the two shared group convolutional layers, the $\text{Conv2d}_{\text{Cls}}$ and $\text{Conv2d}_{\text{Bbox}}$ are the final convolutional layers for the classification and regression tasks, respectively.

GCDetect successfully optimizes the detection head of YOLOv8 by adopting group convolution and feature sharing strategies, which significantly reduces the number of parameters and computational complexity without sacrificing the accuracy and robustness of the model. This optimization not only improves the operation efficiency of the model, but also makes it more suitable for real-time applications, especially in environments with limited computational resources, and significantly improves the inference speed and responsiveness of the model. The structure of GCDetect is shown in Fig. 4.

Figure 4. Structure of the GGCDetect Module

D. Loss: Introducing Inner-MPDIoU Loss Function for Optimization

In this paper, the Inner-MPDIoU loss function is introduced to replace the CIoU loss function in YOLOv8, which solves the deficiency of CIoU in small target detection. Although CIoU considers factors such as overlap area, center of mass distance and aspect ratio, it still has problems in small target localization. Therefore, a more efficient and accurate loss function is needed to optimize the model, especially in the performance of small target detection.

The Inner-MPDIoU loss function combines the Inner-IoU [42], [43] and MPDIoU [44], [45]. The advantages of the two loss functions improve the accuracy of small target detection. Inner-IoU adapts the size of the auxiliary frame by introducing an auxiliary frame and utilizing the hyperparameter ratio to adapt to targets of different scales. The coordinates of the auxiliary frame are calculated as in equation (10):

$$b_l^{\text{gt}} = x_{\text{gt}_c} - \frac{w_{\text{gt}} \cdot \text{ratio}}{2}, b_r^{\text{gt}} = x_{\text{gt}_c} + \frac{w_{\text{gt}} \cdot \text{ratio}}{2}, b_t^{\text{gt}} = y_{\text{gt}_c} - \frac{h_{\text{gt}} \cdot \text{ratio}}{2}, b_b^{\text{gt}} = y_{\text{gt}_c} + \frac{h_{\text{gt}} \cdot \text{ratio}}{2} \quad \#(10)$$

of which $x_{\text{gt}_c}$ and $y_{\text{gt}_c}$ are the center coordinates of the real frame, and $w_{\text{gt}}$ and $h_{\text{gt}}$ are the width and height of the real frame, and ratio is the adjustment factor.

The MPDIoU loss function solves the problem of localization between boxes with the same aspect ratio but different widths and heights. It evaluates the localization accuracy by calculating the distance of key points between the predicted frame and the real frame. The distance is calculated as equation (11):

$$d_1^2 = (b_l^{\text{gt}} - b_l)^2 + (b_t^{\text{gt}} - b_t)^2, \qquad d_2^2 = (b_r^{\text{gt}} - b_r)^2 + (b_b^{\text{gt}} - b_b)^2 \quad \#(11)$$

Combining the two, the Inner-MPDIoU loss function not only considers the centroid, width, height and distance between key points of the real and predicted frames, but also introduces the width and height of the image. This allows the loss function to evaluate the target detection ability of the model more comprehensively, thus providing effective guidance for training and optimization. The specific formulas of Inner-MPDIoU are shown in Eq. (12), Eq. (13), and Eq. (14):

$$\text{inter} = \max\left(0, \min(b_r^{\text{gt}}, b_r) - \max(b_l^{\text{gt}}, b_l)\right) \cdot \max\left(0, \min(b_b^{\text{gt}}, b_b) - \max(b_t^{\text{gt}}, b_t)\right) \quad \#(12)$$

$$\text{union} = (w_{\text{gt}} \cdot h_{\text{gt}} \cdot \text{ratio}^2) + (w \cdot h \cdot \text{ratio}^2) - \text{inter} \quad \#(13)$$

$$\text{InnerMPDIoU} = \frac{\text{inter}}{\text{union}} - \frac{d_1^2}{h_{\text{image}}^2 + w_{\text{image}}^2} - \frac{d_2^2}{h_{\text{image}}^2 + w_{\text{image}}^2} \quad \#(14)$$

where the ratio hyperparameter takes values in the range [0.5, 1.5], the $b_{\text{gt}}$ and $b$ denote the centroids of the true and predicted frames, respectively. By introducing the key point distances $d_1$ and $2$, Inner-MPDIoU can more accurately measure the difference between the predicted frame and the real frame, especially when dealing with targets with the same aspect ratio but different widths and heights. In addition, considering the width and height of the image, the loss function is able to adapt to images with different sizes and ratios, which further improves the ability of small target detection and localization accuracy. The Inner-MPDIoU schematic is shown in Fig. 5.

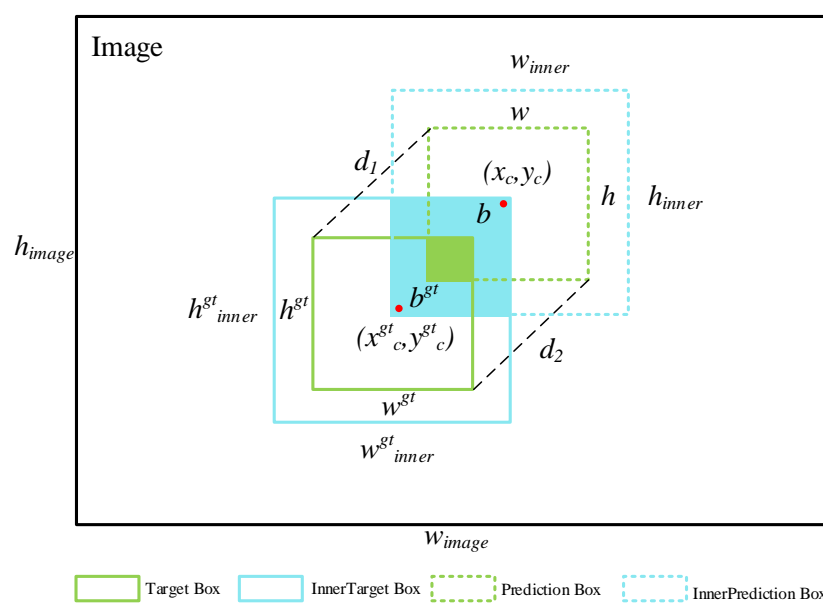

Figure 5. Inner-MPDIoU Schematic Diagram

E. Prune: Implementing Channel Pruning for Model Streamlining

In this paper, we adopt a LAMP (Layer-Adaptive Magnitude-based Pruning) scoring method to prune the layers of YOLOv8. Deploying the trained model on edge-side devices still faces the challenges of poor real-time performance, high computational resource requirements, and high storage occupancy due to the continuous speedup of production. To address these issues, further sub-optimal model compression algorithms are needed, which are capable of converting voluminous models into streamlined models, thus effectively optimizing the models, reducing the deployment cost, and improving the efficiency and performance of the models.

Common model compression algorithms include pruning, quantization, and knowledge distillation, which can be applied individually or in co mbination to obtain better compression results. Choosing the appropriate compression algorithm depends on factors such as the application scenario, target performance requirements, and hardware platforms [46] . Quantization may lead to loss of accuracy, especially in low-bit representations, and requires high hardware support. Knowledge distillation requires training additional student models, which increases computational overhead and time cost. In contrast, pruning reduces model complexity and computational overhead while typically maintaining high accuracy, is easier to deploy on a variety of hardware platforms, and is particularly suitable for the convolutional layer in convolutional neural networks (CNNs). The basic idea is to reduce the number of parameters in the model and the amount of computation by evaluating the importance of each channel and removing those channels that contribute less to the model, while trying to maintain the performance of the model. Channel pruning removes channels that contribute less to the model, so it usually does not significantly affect the performance of the model, and even sometimes improves the generalization ability of the model.

Layer Adaptive Magnitude-based Pruning (LAMP) Scoring by Jaeho Lee et al. [47] is an effective technique for model channel pruning. The core idea of the method is to evaluate the importance of each channel by normalizing the square of the weights. Compared with other methods, the LAMP scoring method does not need to adjust any hyper-parameters, and its computational process is simple and easy to implement, thus making it highly tractable in practical applications.

Place the weight tensor $W$ of the first $u$ LAMP scores for an indicator is defined as equation (15):

$$score(u; W) := \frac{(W[u])^2}{\sum_{v \geq u}(W[v])^2} \#(15)$$

The process of calculating this score consists of squaring the values of each channel in the weight tensor and then summing the squared values of all the channels. By dividing the squared value of each channel by this sum, we are able to obtain the relative importance of each channel. Based on the computed LAMP scores, we can globally cut out the connections with the lowest LAMP scores until the desired global sparsity constraint is satisfied. This process is equivalent to performing an automatically selected minimization process (MP) for hierarchical sparsity.

When the weight $W[u]$ is greater than the square of the weight $W[v]$ the square of the weights, it means that the $W[u]$ that the LAMP score is necessarily higher than $W[v]$ , i.e., equation (16):

$$(W[u])^2 > (W[v])^2 \Rightarrow score(u; W) > score(v; W) \#(16)$$

This logical relationship indicates that connections with larger weights have correspondingly higher LAMP scores. Therefore, during the pruning process, low-weight connections are preferentially removed, which effectively reduces the number of parameters in the network and improves the computational efficiency and performance of the model [48] .

In order to investigate the effect of different pruning rates on model performance, this study conducted ablation experiments by LAMP pruning algorithm. The experimental results show that at a pruning rate of 1.5x, the precision (P) and recall (R) of the model are better than that of the unpruned model, indicating that moderate pruning can effectively improve the model performance. And as the pruning rate increases further (e.g., 2x and above), the precision and recall gradually decrease, especially when the pruning rate reaches 3x, the recall decreases significantly, indicating that excessive pruning may lead to significant degradation of model performance.

In terms of mean accuracy (mAP50) and multi-threshold mean accuracy (mAP50-90), the model's mAP50 and mAP50-90 are both improved at 1.5x pruning, and especially on mAP50-90, the improvement is most significant at 1.5x pruning. As the pruning rate increases, these metrics gradually decline, especially when the pruning rate reaches 3x, the decline is the largest, indicating that over-pruning impairs the model's generalization ability in complex detection tasks.

The pruning operation effectively reduces the number of parameters and computational complexity of the model, and the higher the pruning rate, the significantly lower the storage and computational requirements of the model. Under LAMP 1.5x pruning, the number of parameters of the model is reduced to 49% of that of the unpruned model (671,930 parameters), the computational complexity (GFLOPs) is reduced to 2.4, and the inference speed is increased to 1219 frames per second at a batch size of 32, and the model size is reduced by about 45%. In contrast, 3x pruning, while further reducing the number of parameters and computational complexity, degrades the performance significantly, especially in terms of precision and recall. Thus, while high pruning rates excel in terms of inference speed and storage, the significant performance degradation makes them unsuitable for applications to complex detection tasks.

Taken together, LAMP 1.5x pruning achieves a better balance between improving model performance, reducing model size and improving inference efficiency, and is suitable for finding the best compromise between performance and efficiency. Over-pruning (e.g., 2x and above), on the other hand, may lead to performance degradation, especially in terms of detection accuracy and generalization ability. Therefore, LAMP 1.5x pruning provides an important reference for model optimization in practical applications.

Table 1 Comparison of pruning performance at different pruning rates

|  | NoPrune | Lamp(1.5x) | Lamp(2x) | Lamp (2.5x) | Lamp(3x) |
|---|---|---|---|---|---|
| P | 0.9839 | 0.9891 | 0.9623 | 0.9602 | 0.9578 |
| R | 0.9926 | 0.9941 | 0.9725 | 0.9638 | 0.9345 |
| map50 | 0.9920 | 0.9932 | 0.9701 | 0.9613 | 0.9492 |
| map50-90 | 0.6634 | 0.7518 | 0.7083 | 0.6776 | 0.6267 |
| Parameter | 1375670 | 671930 | 392999 | 272816 | 206094 |
| GFLOPs | 3.7 | 2.4 | 1.8 | 1.4 | 1.2 |
| FPS (bs1) | 126.2 | 122.3 | 126.1 | 129.6 | 132.9 |
| FPS (bs32) | 1032.1 | 1219.4 | 1376.5 | 1476.6 | 1563.5 |
| Model size | 2.9M | 1.6M | 1.0M | 0.8M | 0.7M |

# test

### A. Data sets

For dataset selection, the PCB defect detection dataset PKU-Market-PCB released by the Intelligent Robotics Open Laboratory of Peking University is used in this experiment, which contains a variety of defect types, such as missing hole, mouse bite, open circuit, short, spur, and spurious copper, and so on. Fig. 6 shows the schematic diagram of these six types of PCB defect characteristics. The original dataset contains a to tal of 693 images, and the dataset is expanded to 10668 images through a series of data enhancement methods, including random cropping, random scaling, brightness adjustment, and adding noise, which increases the diversity of the dataset and improves the generalization ability of the model. In order to test the generalization performance of the model, we use the DeepPCB dataset released by the Institute of Image Processing and Pattern Recognition of Shanghai Jiao Tong University, in which all the images are obtained from linear scanning CCDs with a resolution of about 48 pixels per 1 mm, which contains 1500 PCB images, and Fig. 7 shows a schematic diagram of the characteristics of the six types of PCB defects.

The distribution of the PKU-Market-PCB dataset is shown in Table 2, which is divided into training, validation and testing sets in the ratio of 8:1:1 to ensure the comprehensiveness and fairness of the training, validation and testing of the model. Labeling and visualization experiments were conducted for six types of PCB defects, and the results are shown in Fig. 8. Box 1: The amount of data in the training set, showing the number of samples included in each category. Palace 2: Size and number of boxes, showing the size distribution of the bounding boxes in the training set and the corresponding number. Palace 3: Position of the center point relative to the whole image, describing the distribution of the position of the bounding box center point in the image. Grid 4: the height-to-width ratio of the target in the image relative to the whole image, reflecting the distribution of the height-to-width ratio of the target in the training set.

Table 2 PCB data set

| Defect types | Before augmentation images | After augmentation images |
| --- | --- | --- |
| missing hole | 115 | 1832 |
| mouse bite | 115 | 1852 |
| open circuit | 116 | 1740 |
| short | 116 | 1732 |
| spur | 115 | 1752 |
| spurious copper | 116 | 1760 |
| total | 693 | 10668 |

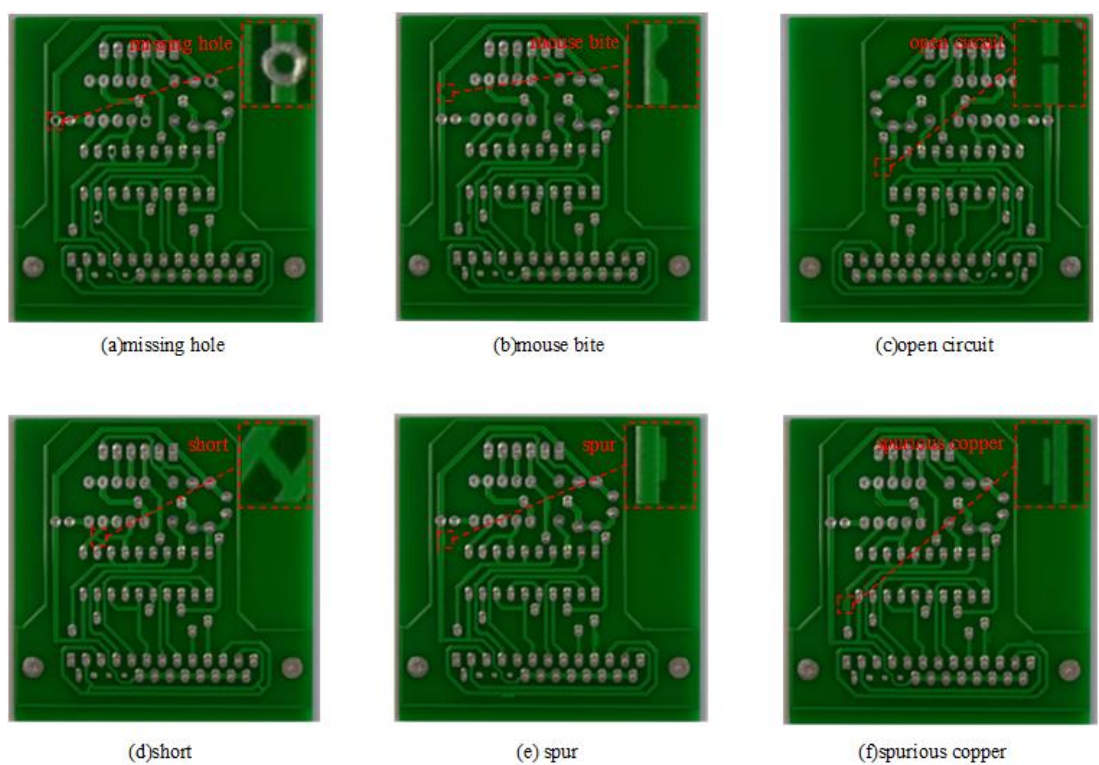

Fig. 6. This figure shows the pair of PKU-Market-PCB defective dataset images

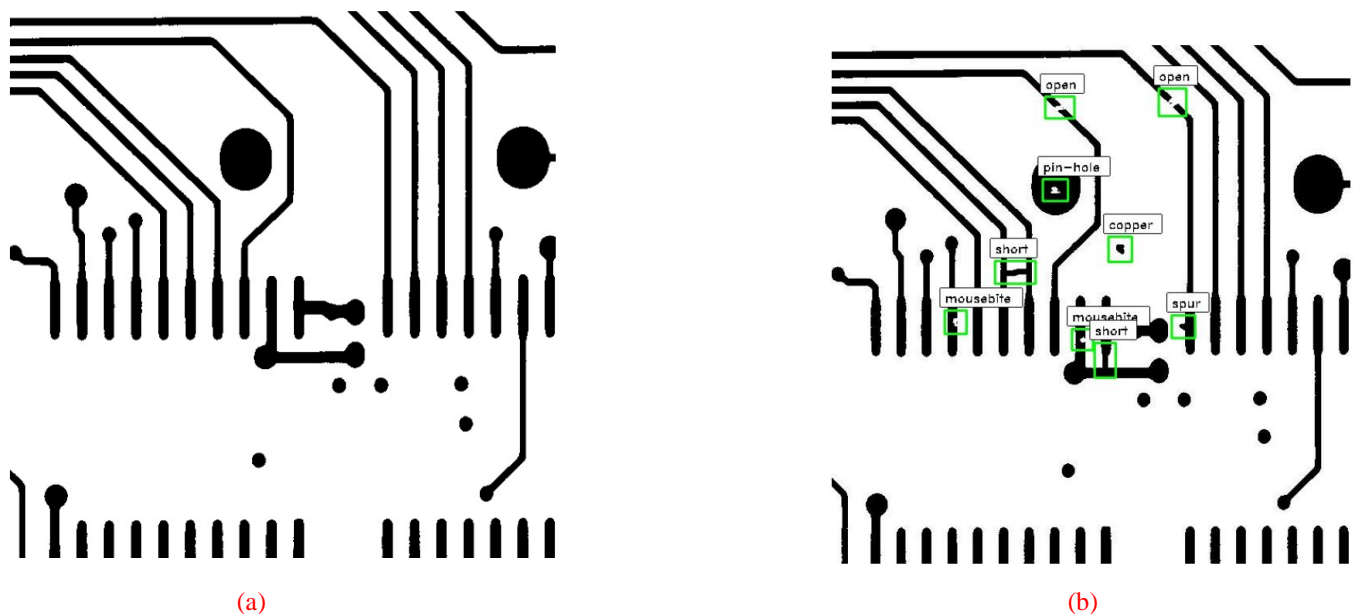

(a)            (b)

This figure shows the pair of (a) a defect-free template image and (b) a defective tested image with annotations of the positions and types of PCB defects in the DeepPCB dataset. defects in the DeepPCB dataset.

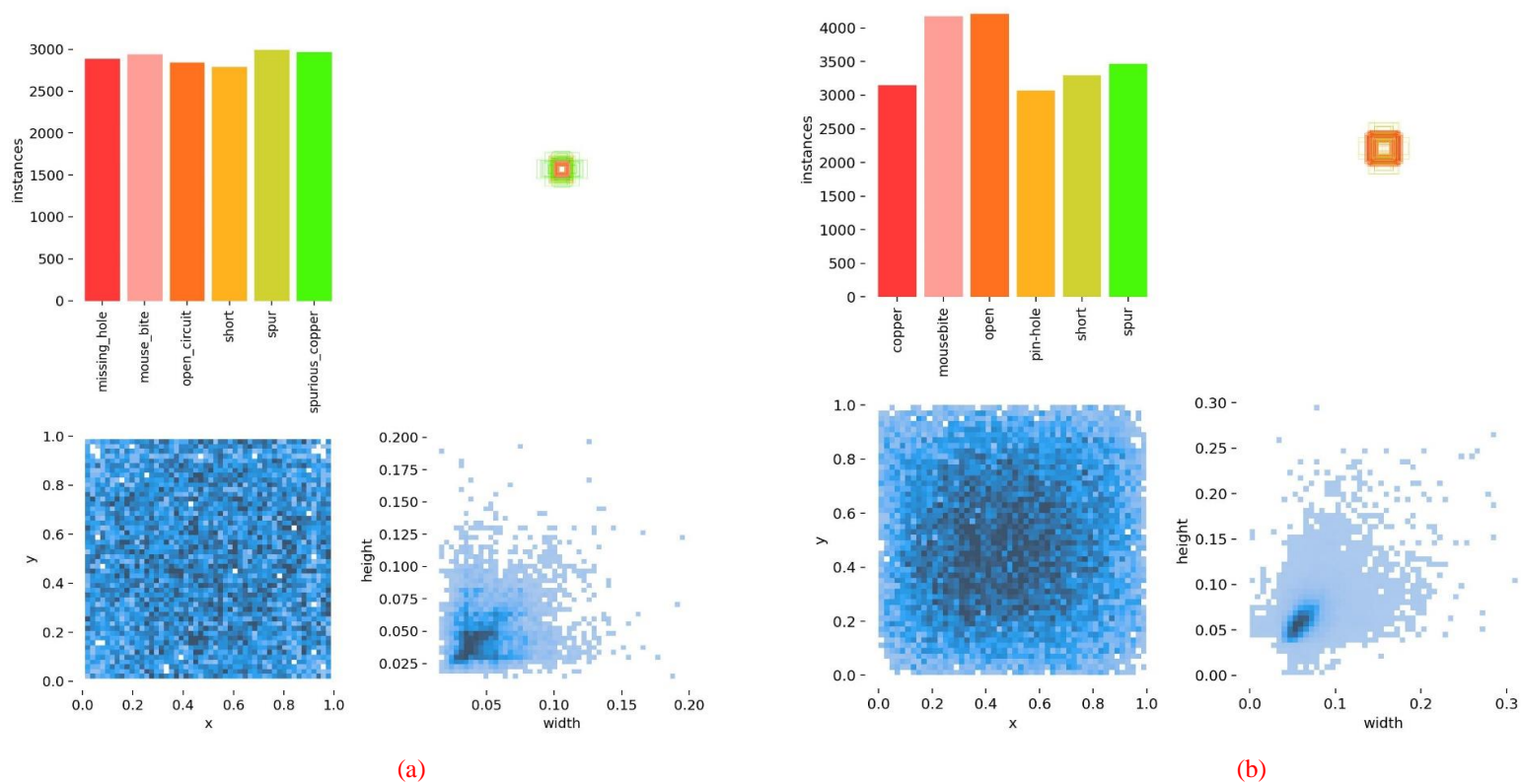

(a)                  (b)

Fig. 8. This figure shows the pair of (a) distribution of PKU-Market-PCB dataset and (b) distribution of DeepPCB dataset.

B. Experimental environment and parameter settings

For the experimental conditions, this study was conducted under Ubuntu 20.04 operating system, using Intel(R) Xeon(R) Silver 4214R CPU @ 2.40GHz as the main processor, equipped with RTX 3080 Ti GPU and utilizing CUDA 11.3 to accelerate the computation, and the deep learning framework was selected as PyTorch 1.10 Experimental environment and parameter settings The experimental environment configuration and parameter settings are shown in Table 3.

Table 3 Experimental environment configuration

| Configuartion Items | Configuration Parameters |
| --- | --- |
| Computer operating system | Ubuntu 20.04 |
| CPU | Intel(R) Xeon(R) Silver 4214R |
| GPUs | NVIDIA RTX 3080 Ti (12GB) |
| Compilation language | Python 3.8 |
| Framework | PyTorch 1.10.0 |
| CUDA | CUDA Version: 11.3 |
| Epochs | 100 |
| Batch size | 16 |
| Image Size | 640× 640 |

C. Evaluation Metrics

Evaluation metrics play a crucial role in deep learning model performance and effectiveness evaluation. To ensure the fairness of the experiments, we perform 100 rounds of training for all networks without using pre-training weights with a batch size of 16. The performance of the models is evaluated using the following metrics: precision(P), recall(R), mean Average Precision(mAP), model size(MB) , Parameter, GFLOPs and FPS.

Evaluation metrics are crucial in industrial scenarios, they are not only a measure of model performance, but also a key indicator for assessing the efficiency of production lines and product quality.Precision and Recall are two basic performance metrics that help us understand the accuracy and comprehensiveness of the model in the inspection process. In industrial production, high Precision means that the model can accurately identify defects and avoid unnecessary false alarms, thus reducing production cost and resource waste. And high recall rate means that the model can effectively capture most of the defects, which reduces the possibility of missed inspection and improves the quality and reliability of the products.

The Precision metric measures the proportion of true positive categories out of all categories predicted to be positive by the model, i.e., the accuracy of the model.Precision is calculated as in equation (17).

$$P = \frac{TP}{TP + FP} \times 100\% \quad \#(17)$$

where TP denotes true cases (i.e., the number of samples that the model correctly predicted as positive categories) and FP denotes false positive cases (i.e., the number of samples that the model incorrectly predicted as positive for samples in negative categories).

The Recall metric measures the proportion of samples successfully predicted by the model to be in the positive category as a percentage of all true positive category samples, i.e., the completeness of the model.Recall is calculated as in equation (18)

$$R = \frac{TP}{TP + FN} \times 100\% \quad \#(18)$$

where FN denotes the number of false negative cases (i.e., the number of samples where the model incorrectly predicts a positive category sample as a negative category).

Ideally, detection models with high values of both p and r perform better. Therefore, it is necessary to combine these two evaluation metrics and assess the performance of the model by combining the average precision (AP).AP is equal to the integral of calculating the p corresponding to each r, which is calculated as in Eq. (19).

$$AP = \int_0^1 P(R)dR \quad \#(19)$$

The mAP (mean accuracy) is a metric that combines Precision and Recall, and it evaluates the performance of a model by calculating and averaging the Precision at different Recall values. In tasks such as target detection, mAP is widely used to measure the model's ability to recognize each category. A higher mAP value indicates that the model has better recognition ability on each category. In industrial production, a high mAP value represents the model's better ability to recognize defects in each category, which can effectively meet the quality control needs of the production line and improve the product qualification rate and customer satisfaction. Its calculation formula is as in equation (20).

$$mAP = \frac{\sum_{i=0}^{n} AP(i)}{n} \#(20)$$

The model size (MB) and the number of parameters (Parameter) reflect the complexity and scale of the model, and usually the smaller the model the easier it is to deploy and reason about. On resource-constrained devices, such as mobile devices and embedded systems, smaller model sizes can save storage space and computational resources, and improve the efficiency of model deployment.

The amount of floating point operations (GFLOPs) is a measure of the computational resources required by the model in the inference phase, which reflects the computational complexity of the model. In edge devices and embedded systems, a lower value of GFLOPs means that the model can run more efficiently in resource-limited environments.

Frames per second (FPS), on the other hand, is a measure of the speed of the model's inference, indicating the number of image frames that the model can process per second. In real-time applications, such as video surveillance and autonomous driving, a higher FPS value means that the model is able to respond and process input data more quickly, improving the real-time and responsiveness of the system.

## Analysis of results

A. Analysis of model training results

Figure 9 shows the loss values, precision, recall, and mAP for each iteration during model training. The training losses include box loss, classification loss, and dynamic feature learning loss, which are represented by train/box_loss, train/cls_loss, and train/dfl_loss, respectively. It can be observed that the loss value of each category fluctuates initially, but gradually decreases and stabilizes as training proceeds. The values of accuracy, recall and mAP gradually increase and stabilize.

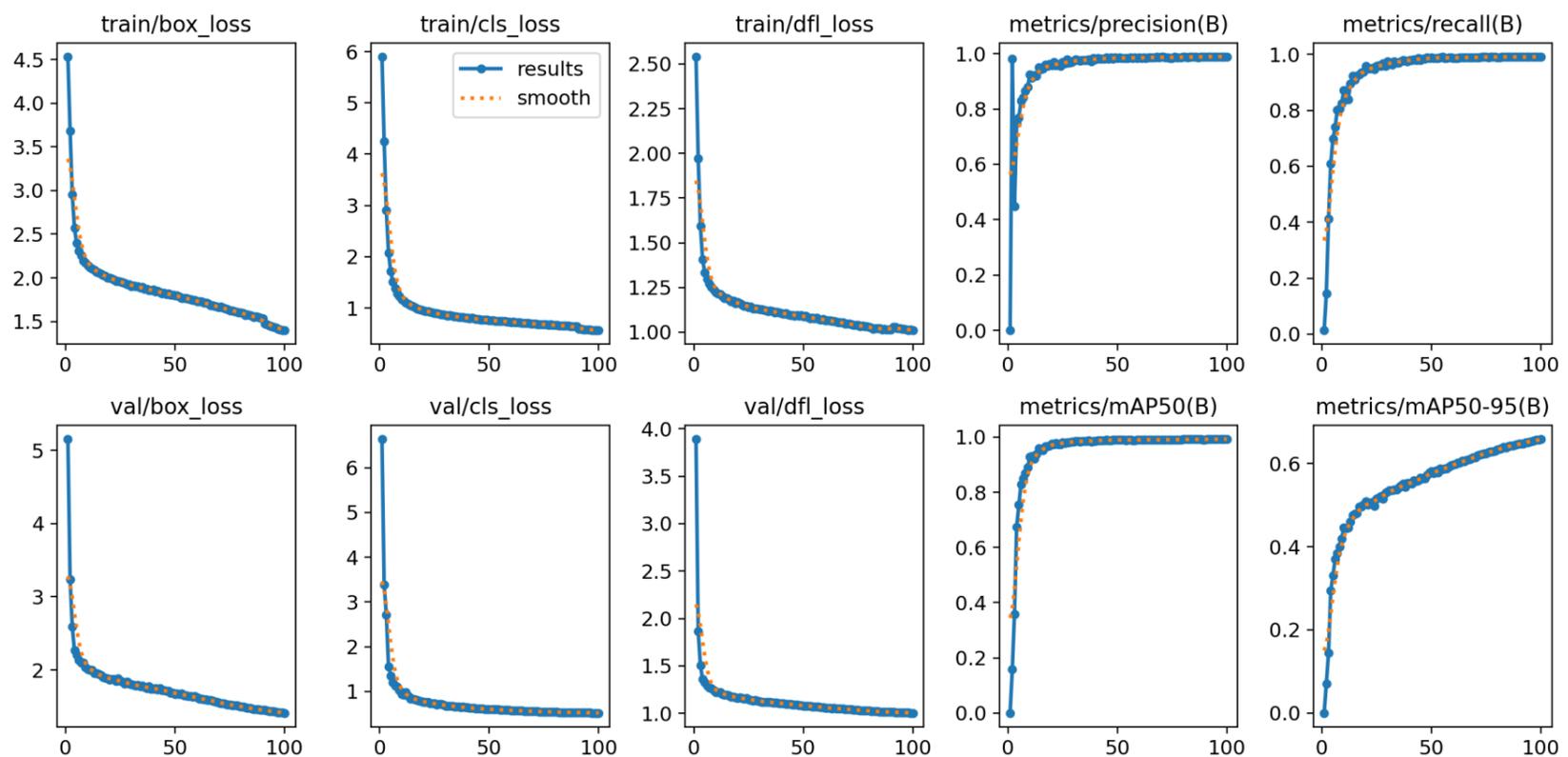

Figure 9. Model metrics during training.

B. Ablation experiments

In this ablation experiment, we conducted a comprehensive evaluation and optimization of different PCB defect detection models, aiming to verify the optimization effect of the models in terms of accuracy and speed. Starting from the initial YOLOv8 model, we gradually introduced a series of improvement methods and conducted an in-depth evaluation and comparison of these improvements, and the results of the ablation experiment data are shown in Table 4. Figure 10 demonstrates the changes in key evaluation metrics between our proposed model and YOLOv8 during the training process. These metrics help us understand the performance of the model and the training progress. The results show that our proposed model outperforms YOLOv8 in all metrics, with significant improvements in mAP, Params, and FPS. In addition, our model starts to stabilize after about 60 training sessions. Compared with YOLOv8, our model shows better performance in terms of training speed and detection accuracy.

First, we integrated GHOST-HGNETV2 into the backbone of YOLOv8. This improvement not only achieves significant compression in terms of the number of parameters and computation volume, but also achieves small improvements in Precision, Recall, mAP0.5 and FPS metrics. The number of parameters is reduced from 3.0M to 2.31M, and the computation volume is reduced from 8.1G to 6.8G, while Precision is improved by 0.3%, Recall is improved by 0.2%, mAP0.5 is improved by 0.4%, and FPS is also slightly improved.

Next, we replaced the original C2F module with C2F-FASTER for the neck network part. Although this improvement results in a slight mAP0.5 drop, it achieves significant reductions in the number of parameters, computation volume, and size. The number of parameters was reduced from 2.31M to 1.92M, the computation volume from 6.8G to 5.5G, and the size from 3.0MB to 2.3MB.

To address the redundancy problem of the head network, we use GCDETECT to replace the original DETECT, and although there is a slight sacrifice in Precision, there is an improvement in Recall and mAP0.5. At the same time, the number of parameters, computation volume and size of the model have been significantly reduced. The number of parameters is reduced to 1.38M, the amount of computation is reduced to 3.7G, and the volume is reduced to 2.9MB.

In order to further improve the model's small target detection ability and localization accuracy, we employed the Inner-MPDIoU loss function to train the model. This improvement not only achieves a certain degree of enhancement on Precision, Recall and mAP0.5, but also has better generalization ability when dealing with PCB defect detection tasks.

Finally, we performed channel pruning on the improved model. This process greatly reduces the number of parameters, computation and volume of the model, while maintaining the advantages of the model in Precision, Recall, mAP and FPS. The number of parameters of the improved model is only 0.67M, the amount of computation is only 2.4G, and the model volume is only 1.6M.

In terms of accuracy, our model outperforms the YOLOv8 model in Precision, Recall, mAP0.5 and mAP0.5:0.95 metrics. Specifically, our model achieves 98.91% in Precision, which exceeds the YOLOv8 model by 0.47%; Recall reaches 99.41%, which is 0.36 percentage points higher than the YOLOv8 model; mAP0.5 reaches 99.32%, which is 0.16% higher; and mAP0.5:0.95 reaches 75.18%, which is 75% higher than the YOLOv8 model. mAP0.5 reached 99.32%, an increase of 0.16%, while mAP0.5:0.95 reached 75.18%, an increase of 10.13% over the YOLOv8 model.

In terms of speed, our model also achieves a significant improvement. In terms of detection speed, our model achieves 129.3 FPS, which is an improvement of 10.2 FPS compared to the 119.1 FPS of the YOLOv8 model, while in the case of batch size of 32, the FPS of our model improves from 1030.1 to 1219.4 FPS, which further strengthens the advantage of our model in real-time.

By introducing a series of optimization methods such as Ghost-HGNetV2, C2f-Faster, GCDetect, Inner-MPDIoU, and channel pruning, our model shows significant advantages in the PCB defect detection task. Not only does it outperform the YOLOv8 model in terms of Precision, Recall, mAP0.5 and mAP0.5:0.95, but it also achieves significant improvement in detection speed.

These results fully demonstrate that our method outperforms the traditional YOLOv8 model in terms of accuracy and speed, and provides an important reference value for the practical application of PCB defect detection tasks.

Table 4 Ablation experiments

| A | B | C | D | E | P (%) | R(%) | mAP0.5 (%) | mAP0.5:0.95 (%) | Params(M) | FLOPs (G) | FPS (bs1) | FPS (bs32) | Size(MB) |
|---|---|---|---|---|-------|------|------------|-----------------|-----------|-----------|-----------|------------|----------|
|   |   |   |   |   | 98.44 | 99.05 | 99.16 | 65.05 | 3.00 | 8.10 | 119.10 | 1030.10 | 6.00 |
| ✓ |   |   |   |   | 98.60 | 99.15 | 99.21 | 64.35 | 2.31 | 6.80 | 120.80 | 1034.30 | 4.70 |
| ✓ | ✓ |   |   |   | 98.68 | 98.78 | 98.98 | 63.28 | 1.97 | 6.10 | 118.90 | 1011.00 | 4.00 |
| ✓ | ✓ | ✓ |   |   | 98.25 | 98.85 | 99.06 | 63.51 | 1.38 | 3.70 | 126.70 | 1035.20 | 2.90 |
| ✓ | ✓ | ✓ | ✓ |   | 98.39 | 99.26 | 99.20 | 66.34 | 1.38 | 3.70 | 126.70 | 1035.20 | 2.90 |
| ✓ | ✓ | ✓ | ✓ | ✓ | 98.91 | 99.41 | 99.32 | 75.18 | 0.67 | 2.40 | 129.30 | 1219.40 | 1.60 |

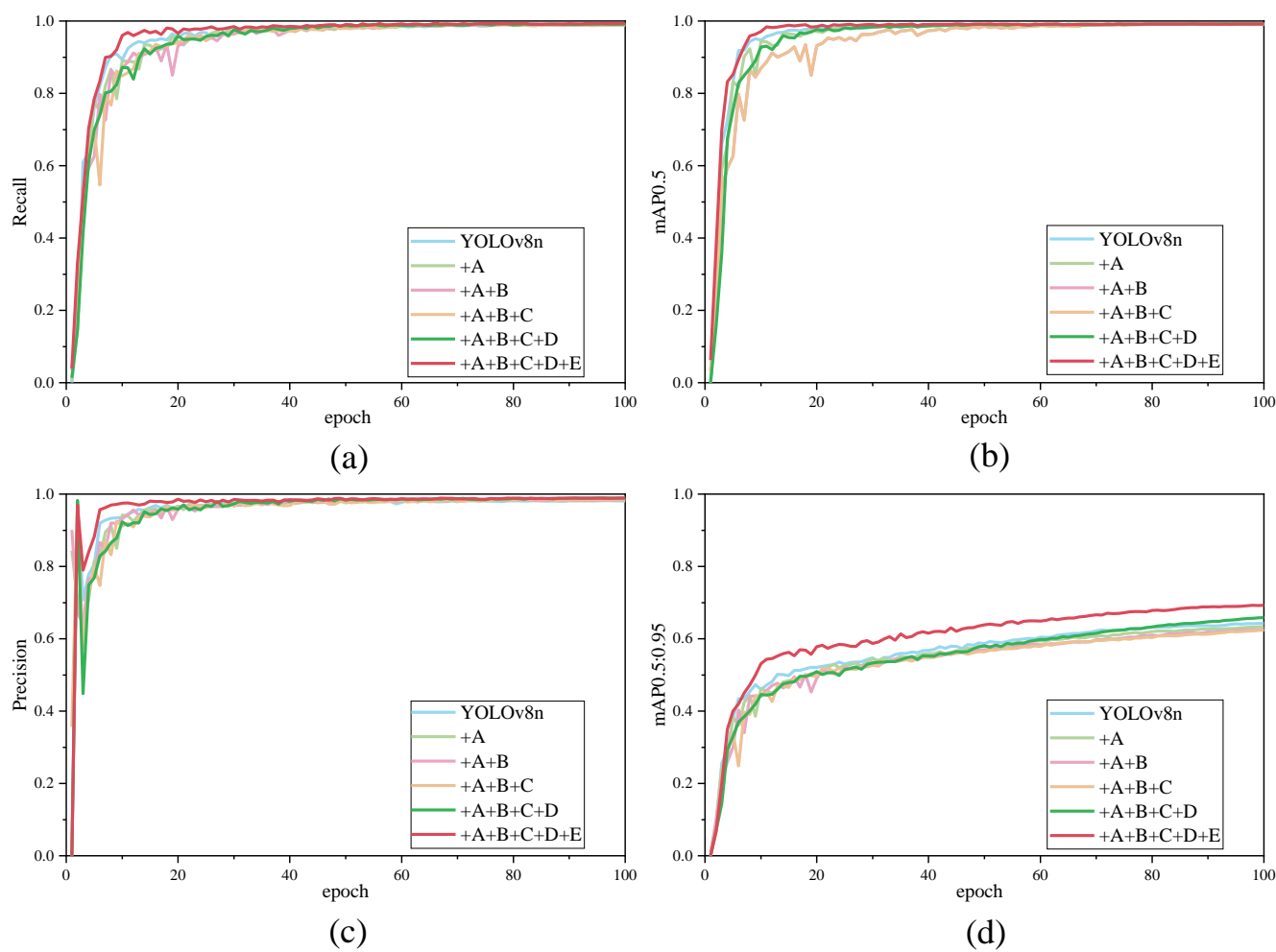

Fig. 10 Graph of ablation experiment results

B. Comparison of algorithms

In this paper, we conduct a comprehensive comparative analysis of the performance of our proposed network with that of traditional target recognition networks in the PCB defect detection process, aiming to highlight the superiority of our model. In this study, we compare in detail the differences between our model and other commonly used target detection algorithms using accuracy and speed as evaluation metrics, and the data from the comparison experiments are shown in Table 5.

First of all, our model shows obvious advantages in Precision and Recall. Compared with traditional algorithms such as SSD and Faster-RCNN, our model not only achieves higher values in Precision and Recall, but also improves significantly in mAP0.5 and mAP0.5:0.95 metrics. Particularly, in mAP0.5, our method improves the performance by 6.1% and 6.9% compared to SSD and Faster-RCNN, respectively, showing the significant improvement of our model in detection accuracy.

Further, our model also presents a significant advantage when comparing with different versions of the YOLO family. Relative to YOLOv3-tiny, YOLOv4-tiny, YOLOv5n, YOLOv6n, and YOLOv7-tiny, our model outperforms the mAP0.5 values by 3.3%, 9.7%, 1.2%, 0.6%, and 3.1%, respectively. This indicates that our method not only provides higher accuracy in PCB defect detection tasks, but also has stronger applicability and generalization ability.

The performance advantage of our model is even more significant in terms of mAP0.5:0.95. Relative to SSD and Faster-RCNN, our method outperforms mAP0.5:0.95 by 28.5% and 29.9%, respectively, whereas our model outperforms the mAP0.5:0.95 value by 27.8%, 40.2%, 17.0%, 18.1%, and 26.5%, respectively, relative to the different versions of the YOLO series. This result further strengthens the strong performance of our model in terms of detection accuracy.

In addition to the advantages in detection accuracy, our model also excels in model detection speed, achieving a speed of 129.3 FPS. This speed is much higher than traditional algorithms such as SSD and Faster-RCNN, demonstrating the great potential and advantages of our model in real-time application scenarios.

Our model in the field of PCB defect detection not only excels in accuracy, but also has a significant advantage in speed. This series of advantageous results fully demonstrates the powerful performance of our model in practical applications, providing a more reliable and efficient solution for PCB defect detection tasks, which has important practical application value and promotion significance.

Table 5 Algorithm comparison experiment

| Experiments | P (%) | R(%) | mAP0.5 (%) | mAP0.5:0.95 (%) | Params(M) | FLOPs (G) | FPS |
|---|---|---|---|---|---|---|---|
| SSD300 | 91.31 | 90.13 | 93.20 | 46.61 | 24.40 | 30.70 | 47.30 |
| Faster-RCNN | 89.12 | 95.80 | 92.41 | 45.21 | 137.10 | 370.20 | 23.60 |
| Yolov3-tiny | 95.04 | 93.13 | 96.09 | 47.32 | 8.70 | 12.90 | 64.60 |
| Yolov4-tiny | 95.81 | 79.65 | 89.66 | 34.91 | 6.10 | 16.50 | 65.20 |
| Yolov5n | 96.72 | 96.33 | 98.14 | 58.19 | 1.80 | 4.20 | 110.70 |
| Yolov6n | 98.16 | 95.52 | 98.73 | 57.01 | 4.70 | 11.40 | 90.30 |
| Yolov7-tiny | 95.98 | 91.71 | 96.22 | 48.63 | 6.00 | 13.10 | 82.50 |
| Yolov9-c | 98.38 | 99.31 | 99.32 | 68.70 | 25.30 | 102.10 | 33.30 |
| Yolov8n | 98.44 | 99.05 | 99.16 | 65.05 | 3.00 | 8.10 | 119.10 |
| Our model | 98.91 | 99.41 | 99.32 | 75.18 | 0.70 | 2.40 | 129.30 |

C. Generalization experiments

In this experiment, we evaluate the generalization performance of the original YOLOv8 model and the improved YOLOv8 model with DeepPCB as the experimental dataset. The experiment mainly compares the performance of the two models in terms of Detection Precision (Precision, P), Recall (Recall, R), Mean Average Precision (mAP0.5 and mAP0.5:0.95), Number of Model Parameters (Params), Computation (FLOPs), and Model Size (Size). (mAP0.5 and mAP0.5:0.95), the number of model parameters (Params), the amount of computation (FLOPs), and the model size (Size).

The experimental results show that our model outperforms the original YOLOv8n model in several key performance indicators. In terms of detection accuracy, the P-value of the improved model reaches 95.18%, which is significantly improved compared to the 92.91% of the original model; at the same time, the mAP0.5 and mAP0.5:0.95 of the improved model reach 97.29% and 70.96%, respectively, which are 1.12% and 8.47% higher compared to the 96.17% and 62.49% of the original model. These results show that the improved model exhibits better generalization ability in both accuracy and average precision.

In addition, the improved model also shows a small improvement in recall with an R-value of 92.23% compared to 91.65% for the original model. Although the improvement is small, it still shows the optimization of the improved model in terms of target detection rate. Most importantly, the improved model shows a significant advantage in terms of resource consumption. Its parameter amount is reduced to 0.67M, which is much smaller than the 3.00M of the original model, a reduction of 77.67%; at the same time, the FLOPs are also significantly reduced from 8.10G to 2.4G of the original model, a reduction of 70.37%. This indicates that the improved model greatly reduces the computational complexity and storage requirements while maintaining high detection accuracy.

In terms of model size, the improved model is only 1.6MB, which is significantly smaller than the 6.00MB of the original YOLOv8 model, a reduction of 73.33%. This result further indicates that the improved model has better deployment potential in resource-constrained environments, especially for embedded devices or other scenarios that require lightweight models.

The experimental results of the improved YOLOv8 model on the DeepPCB dataset are shown in Table 6, which significantly outperforms the original YOLOv8 model in terms of performance metrics such as detection accuracy and average accuracy, and at the same time, it is significantly optimized in terms of the number of parameters, computation volume and model size, achieving a good balance between performance and resource consumption. This makes the improved model have stronger generalization ability and deployment flexibility in practical applications.

Table 6 Results of generalization experiments on DeepPCB dataset

|  | P (%) | R(%) | mAP0.5 (%) | mAP0.5:0.95 (%) | Params(M) | FLOPs (G) | Size(MB) |
|---|---|---|---|---|---|---|---|
| YOLOv8n | 92.91 | 91.65 | 96.17 | 62.49 | 3.00 | 8.10 | 6.00 |
| Our | 95.18 | 92.23 | 97.29 | 70.96 | 0.67 | 2.40 | 1.60 |

D. discuss the limitation

The experimental results (Fig. 11) show that PCB defect detection using YOLOv8 and our improved model exhibits significant advantages in small target detection. The types of defects detected include missing holes, rat bites, open circuits, short circuits, spurs, and false copper. Each defect instance is labeled by a bounding box and is accompanied by a confidence score, with higher confidence indicating a higher likelihood of the defect occurring. Compared with the original YOLOv8 model, our improved method performs more prominently on small target detection.

By optimizing the network structure and introducing a new loss function, our model not only improves the detection accuracy, but also significantly improves the recognition of small targets. In complex contexts, our model is able to detect small defects on PCBs quickly and accurately with higher confidence scores. This enhancement means that our model has higher reliability and accuracy in locating and identifying small defects.

Nevertheless, the model still has some limitations in practical applications. For example, the detection performance may be degraded when dealing with blurred images, which may be affected by the shooting equipment and environmental conditions in real production. In addition, for defect types with complex shapes or blurred edges, the model's bounding box labeling may not be accurate enough, which may affect the stability of detection. These issues suggest that we need to optimize the data enhancement strategy in the future to further improve the model's ability to adapt to complex scenes and different defect types, in order to meet more practical application requirements.

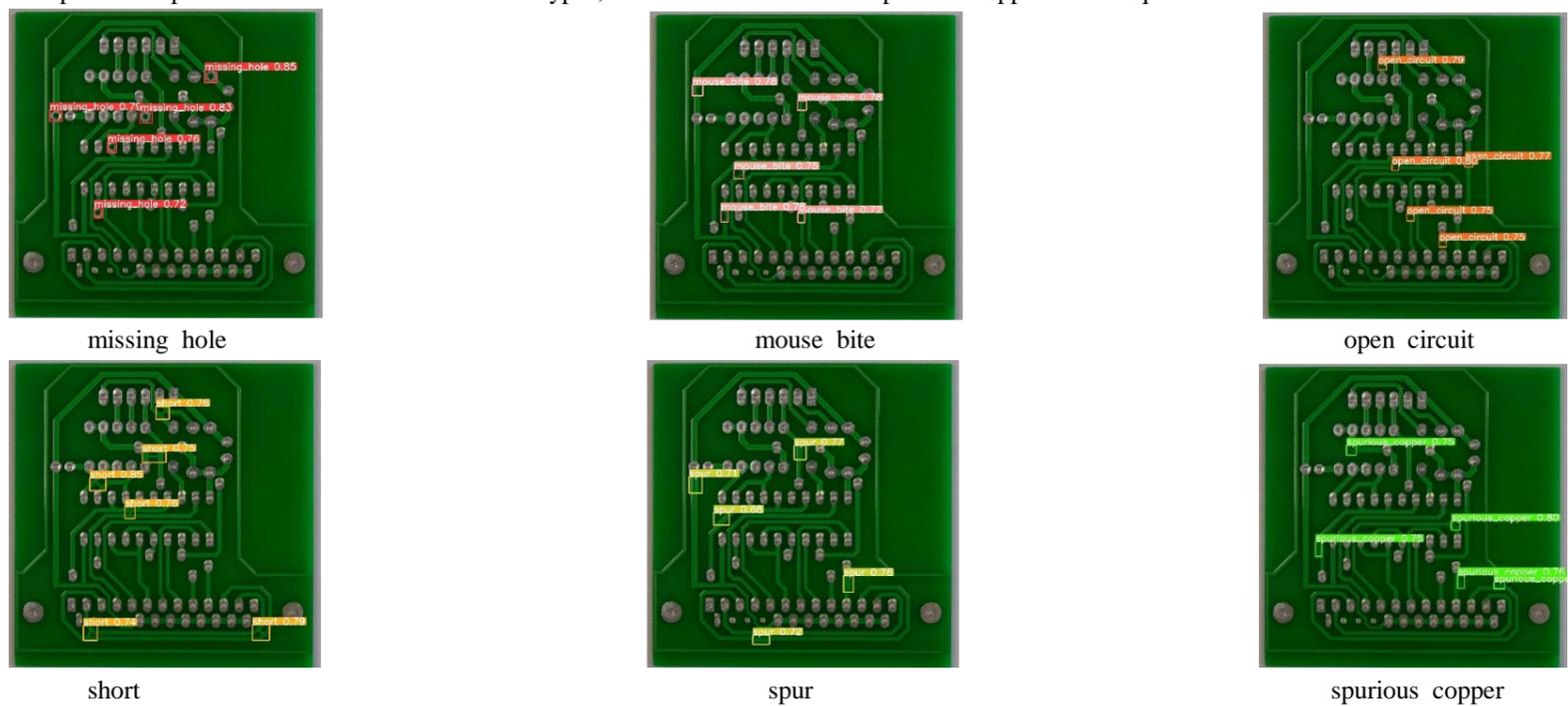

(a) Yolov8

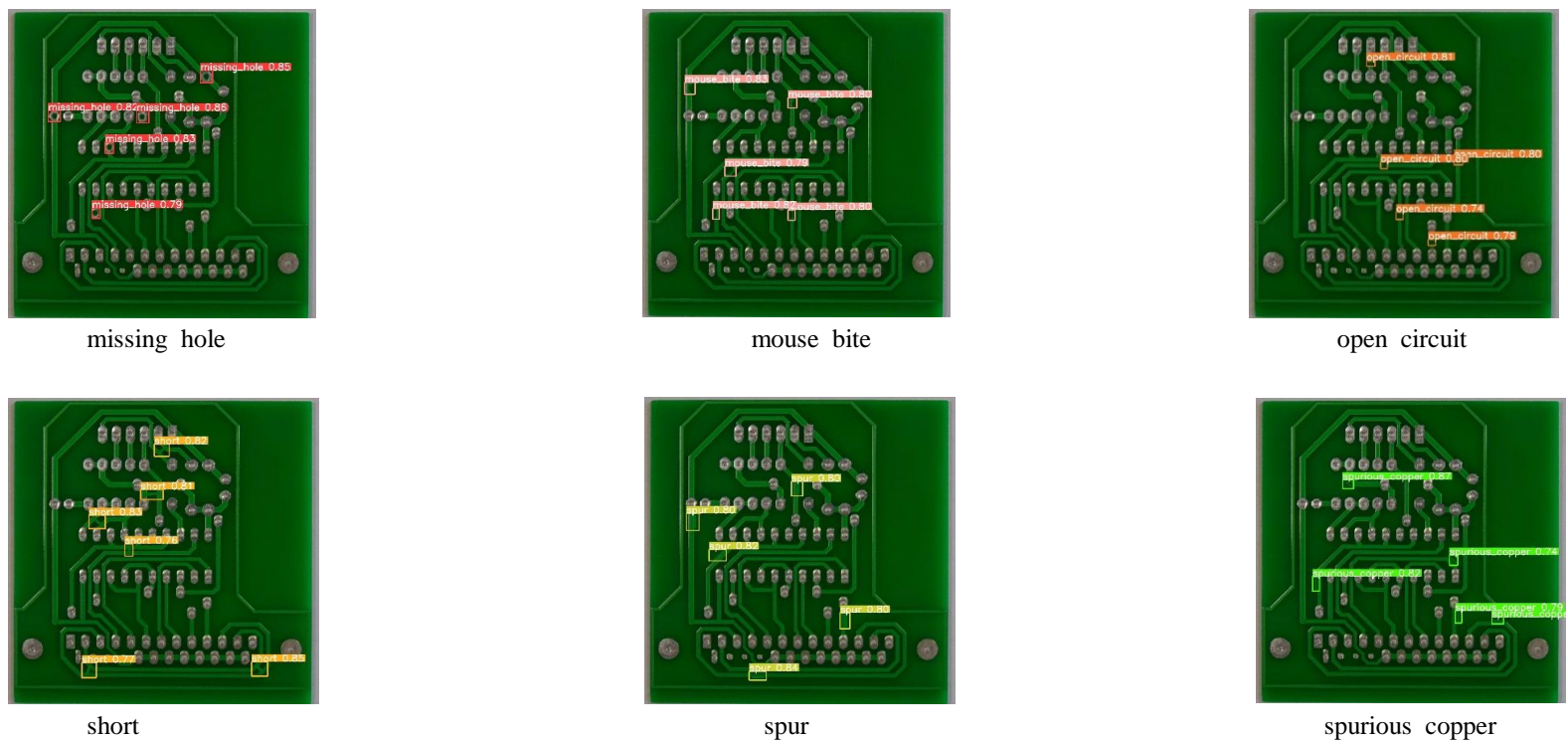

(b) Our Model

Fig. 11 Detection effect diagram for PKU-Market-PCB dataset

## summarize

Compared with the current popular target detection network models, this paper introduces a lightweight YOLOv8 model specialized for detecting PCB defects. In this paper, we first improve the Ghost-HGNetV2 backbone with multilevel feature extraction strategy, which solves the problem of feature information loss in the backbone network , and enhances the model's ability to detect small objects by utilizing multiscale information. Meanwhile, a lightweight feature fusion C2f-Faster module and GCDetect detection head are used to reduce the number of network parameters and computational complexity and improve the computational efficiency of the network while ensuring the detection accuracy. Subsequently, a novel Inner-MPDIoU loss function is used to guide the model to accurately predict the location of the bounding box, which effectively improves the small target detection performance and localization accuracy of the model. Finally, in order to further compress the model volume, we performed LAMP channel pruning on the model to explore having an optimized pruning rate and substantially compress the model volume.

The experimental results clearly show that the improved target detection model in this study has significant advantages on the PCB dataset.

The model performs well in terms of detection accuracy and speed, meeting the requirements for lightweight deployment. In addition to meeting the requirements of high detection accuracy, the compact design of the model makes it suitable for embedded deployment, providing strong support for practical applications.

In future work, we will further optimize the model to enhance its robustness and efficiency in complex environments. Specifically, explore more innovative network architectures, such as adaptive feature fusion and efficient feature selection mechanisms, to better utilize multi-scale information and further reduce model size. Investigate effective incremental learning algorithms that enable the model to quickly learn new defect features and adapt to new detection requirements without retraining the entire network, while maintaining the lightweight nature of the model.